\documentclass[10pt]{article} 

\usepackage[accepted]{rlc}

\usepackage{amssymb}            
\usepackage{mathtools}          
\usepackage{mathrsfs}           
\mathtoolsset{showonlyrefs}     
\usepackage{graphicx}           
\usepackage{subcaption}         
\usepackage[space]{grffile}     
\usepackage{url}                
\usepackage{xcolor}
\usepackage{latexcolors}
\usepackage{enumerate, amsmath, amsthm, amssymb, algorithm, algorithmic, pifont, fullpage, csquotes, dashrule, tikz, bbm, booktabs, bm, verbatim, url, mathrsfs}
\usepackage{dsfont}
\usepackage{wrapfig}

\definecolor{dblue}{RGB}{98, 140, 190}
\definecolor{dlblue}{RGB}{216, 235, 255}
\definecolor{dgreen}{RGB}{124, 155, 127}
\definecolor{dpink}{RGB}{207, 166, 208}
\definecolor{dyellow}{RGB}{255, 248, 199}
\definecolor{dgray}{RGB}{46, 49, 49}

\newcommand{\durl}[1]{\textcolor{dblue}{\underline{\url{#1}}}}

\newcommand{\mc}[1]{\mathcal{#1}}

\newcommand{\bE}{\mathbb{E}}

\newcommand{\bR}{\mathbb{R}}
\newcommand{\bP}{\mathbb{P}}

\newcommand{\bI}{\mathbb{I}}

\newcommand{\bZ}{\mathbb{Z}}

\DeclareMathOperator*{\argmax}{arg\,max}

\newcounter{DaveDefCounter}
\newtheorem{conjecture}{Conjecture}

\newtheorem{example}{Example}

\newtheorem{theorem}{Theorem}

\def\actions{\mathcal{A}}

\def\1{\mathbf{1}}
\def\E{\mathbb{E}}

\def\Pr{\mathbb{P}}

\def\1{\mathbf{1}}

\def\hat{\widehat}

\newif\ifsubmit
\submittrue
\ifsubmit
\newcommand{\dnote}[1]{}
\newcommand{\wnote}[1]{}
\newcommand{\bnote}[1]{}
\else
\newcommand{\dnote}[1]{\textcolor{blue}{Dilip: #1}}
\newcommand{\wnote}[1]{\textcolor{turquoise}{Wanqiao: #1}}
\newcommand{\bnote}[1]{\textcolor{violet}{Ben: #1}}
\fi

\usetikzlibrary{automata} 
\usetikzlibrary{positioning} 
\usetikzlibrary{arrows} 
\tikzset{node distance=2.5cm, 
every state/.style={ 
semithick,
fill=gray!10},
initial text={}, 
double distance=2pt, 
every edge/.style={ 
draw,
->,>=stealth', 
auto,
semithick}}
\let\epsilon\varepsilon

\title{Exploration Unbound}

\author{Dilip Arumugam$^*$  \\
    \texttt{dilip@cs.stanford.edu} \\
    Department of Computer Science\\
    Stanford University
    \And
    Wanqiao Xu$^*$ \\
   \texttt{wanqiaoxu@stanford.edu}\\
    Department of Management Science \& Engineering\\
    Stanford University
    \And
    Benjamin Van Roy \\
    \texttt{bvr@stanford.edu}\\
    Department of Electrical Engineering\\
    Department of Management Science \& Engineering\\
    Stanford University}

\def\actions{\mathcal{A}}

\def\E{\mathbb{E}}

\def\Pr{\mathbb{P}}

\def\1{\mathbf{1}}

\newcommand{\gdot}{discounted-overtaking optimal}

\begin{document}
\def\thefootnote{*}
\footnotetext{Equal contribution}
\maketitle

\begin{abstract}
A sequential decision-making agent balances between exploring to gain new knowledge about an environment and exploiting current knowledge to maximize immediate reward. For environments studied in the traditional literature, optimal decisions gravitate over time toward exploitation as the agent accumulates sufficient knowledge and the benefits of further exploration vanish.  What if, however, the environment offers an unlimited amount of useful knowledge and there is large benefit to further exploration no matter how much the agent has learned?  We offer a simple, quintessential example of such a complex environment.  In this environment, rewards are unbounded and an agent can always increase the rate at which rewards accumulate by exploring to learn more.  Consequently, an optimal agent forever maintains a propensity to explore.
\end{abstract}

\section{Introduction}
\label{sec:intro}

Consider an unscrupulous geometry teacher and a persistent student eager to learn the digits of the mathematical constant $\pi \approx 3.1415926535$. On each day, the student approaches the teacher and may provide any arbitrarily-long sequence of digits. For any $k$-digit sequence they provide, the teacher simply checks to see if these exactly match the first $k$ digits of $\pi$ or not. If there is such an exact $k$-digit match, the teacher awards the student $r(k)$ dollars; otherwise, for anything less than a perfect match, the teacher charges the student $c(k)$ dollars. The teacher sets $r(k)$ and $c(k)$ to be increasing functions of the number of digits and, therefore, guessing longer digit sequences offers more potential upside but also increased risk from erroneous guesses.

Suppose rewards are bounded and the student wishes to maximize expected discounted return.  Then, if on any day, it is optimal to exploit current knowledge by choosing $k$ digits of $\pi$ the student has learned thus far, it will be optimal to choose those same $k$ digits of $\pi$ on every subsequent day.  This is because the theory of sequential decision-making problems establishes that, when rewards are bounded, there is an optimal policy that maps each possible state of knowledge to a single action.  In our example, the student's state of knowledge remains unchanged by the exploitative choice.  This leads to indefinite daily repetition of the same knowledge state and same optimal action.

On the other hand, if rewards are unbounded, there may always be substantial value in exploring to learn more.  In this circumstance, the aforementioned behavior, where on some date the student exploits and continues to do so indefinitely, can fall far short of optimal.  Instead, an optimal policy may have to randomize between exploration and exploitation, without ever tapering the intensity of the former.  This randomization must strike a delicate balance as exploration can be costly and exploitation is required to recoup those costs.

The preceding example represents but one instance of a much broader class of complex environments, for which policies that taper exploration forgo value.  In this paper, we will introduce and analyze a representative instance to offer insights into optimal behavior. Beyond this paper, these complex environments encapsulate an increasingly ubiquitous exploration challenge that modern decision-making systems confront at a much grander scale, comparable to that of the World Wide Web~\citep{shi2017world,toyama2021androidenv,stiennon2020learning,ouyang2022training,yao2022webshop,dwaracherla2024efficient}. Consider a fixed user prompt submitted to a large language model (LLM). To any current response produced by the LLM, the versatility of natural language affords opportunities to explore a vast space of alternative responses and identify more valuable ones.  Clearly, pure exploration would prove very costly because users would largely receive less-desirable, experimental responses.  At the same time, our results suggest that tapering exploration would be sub-optimal as repeating any current response could be improved with further exploration and learning.

The paper proceeds as follows: we formulate a general class of bandit-learning problems in Section \ref{sec:prob_form}, introduce in Section \ref{sec:didactic} our complex bandit environment which admits perpetual improvement that demands eternal exploration, and conclude with discussion as well as future outlook in Section \ref{sec:disc}. For ease of readability, all technical proofs have been relegated to the appendix.

\section{Problem Formulation}
\label{sec:prob_form}

We consider a {\bf bandit environment}~\citep{lattimore2020bandit} with a (possibly infinite) action set $\actions$ and stochastic reward process $\{R_t\}_{t \in \bZ_{>0}}$. Here, $\{R_t\}_{t \in \bZ_{>0}}$ is an exchangeable sequence of random vectors, each with one component per action; each vector $R_t$ assigns a scalar reward $R_{t,a}$ to each action $a$.  At each time $t \in \mathbb{Z}_{\ge0}$, an action $A_t$ is executed and generates a scalar reward $R_{t+1, A_t}$.  If there exists an $R\in\mathbb{R}$ such that $\E[R_{t,a}] < R$ for each time $t$ and action $a$, we say that rewards are bounded. Otherwise, we say that rewards are unbounded. 

By de Finetti's Theorem~\citep{de1937foresight}, exchangeability immediately implies the existence of a random variable $\theta$, representing the unknown bandit environment, conditioned upon which the sequence $\{R_t\}_{t \in \bZ_{>0}}$ is iid.  A common example may clarify our formulation; consider the Bernoulli bandit with unknown success probabilities $\theta \in [0,1]^\actions$ and observe that, conditioned on $\theta$, the rewards $\{R_t\}_{t \in \bZ_{>0}}$ are iid.  Hence, $\{R_t\}_{t \in \bZ_{>0}}$ is exchangeable.  The initial distribution of $\theta$ expresses prior beliefs.  At time $t$, there is a history $H_t = (A_0,R_{1,A_0}, A_2, \ldots, A_{t-1}, R_{t,A_{t-1}})$ of actions and realized rewards, and posterior beliefs are expressed by the distribution of $\theta$ conditioned on $H_t$.

A (stationary) {\bf policy} $\pi$ is a mapping from histories to action probabilities.  In particular, for any history $h$ and action $a$, $\pi(a \mid h)$ is the probability assigned to executing action $a$ upon observing history $h$.  Hence, if an agent applies a policy $\pi$, each action $A_t$ is sampled from $\pi(\cdot \mid H_t)$. To frame a notion of optimality across policies, we first define the expected finite-horizon discounted return of a policy $\pi$:
$$V_\pi^{\gamma,T} = \E\left[\sum_{t=0}^{T-1} \gamma^t R_{t+1,A_t}\right].$$
Here, $\gamma \in [0,1]$ is a discount factor and $T \in \bZ_{>0}$ is a time horizon.  Note that this expectation integrates over uncertainty not only in $\theta$ but also in rewards conditioned on $\theta$.

A policy $\pi$ is said to be {\bf discounted-overtaking optimal} if, for all policies $\pi'$,
$$\liminf_{T \rightarrow \infty} (V^{\gamma, T}_\pi - V^{\gamma, T}_{\pi'}) \geq 0.$$
If $\gamma = 1$, this objective corresponds to the standard notion of overtaking optimality (see Section 5.4.2 of \citep{puterman1994markov}). 
If rewards are bounded, a policy is discounted-overtaking optimal if and only if it maximizes the expected discounted return $V_\pi^\gamma \equiv V_\pi^{\gamma,\infty}$, which is finite.  On the other hand, if rewards are unbounded, $V_\pi^\gamma$ can be infinite for many policies.  The more general notion of discounted-overtaking optimality affords more nuanced comparisons among policies that attain infinite discounted return.  Intuitively, if there is a set of policies that attain infinite expected discounted return, only those that more quickly accumulate discounted rewards can be discounted-overtaking optimal.

While discounted-overtaking optimality offers an unambiguous criteria for a best policy, a \gdot{} policy may not always exist.  To facilitate comparing policies without knowledge of a \gdot{} policy, we define the expected finite-horizon regret of a policy $\pi$ relative to a reference policy $\mu$:
$$
\mathrm{Regret}^{T}_{\pi,\mu} = \E_\pi \left[\sum_{t=0}^{T-1} R_{t+1,A_t} \right] - \E_\mu \left[\sum_{t=0}^{T-1} R_{t+1,A_t} \right].
$$

\section{Necessity of Randomized Exploration}
\label{sec:didactic}

In this section, we define and analyze a complex environment that will be our main object of study for the remainder of the paper.

\begin{example}
We define a bandit with an action set $\actions = \bigcup\limits_{k=0}^\infty \bZ_{> 0}^k$.  Hence, each action $a \in \actions$ is a positive-integer-valued tuple $(a_1,\ldots,a_k)$ of some arbitrary length $k \geq 0$. The stochastic process of reward vectors $\{R_{t+1}\}_{t \in \bZ_{\geq 0}}$ is parameterized by fixed, known scalars $\alpha > 1$ and $\tau \in (1,\frac{\gamma(\alpha-1)}{2(1-\gamma)})$, and an unknown positive-integer sequence $a^* = (a_1^*, a_2^*, a_3^*, \ldots)$ such that, for each $a \in \bZ_{>0}^k$,
$$R_{t+1,a} = \left\{\begin{array}{ll}
\alpha^k \qquad & \mathrm{if }\ a = a^*_{1:k} \\
- \frac{\alpha+1}{\tau-1}\alpha^{k-1} & \mathrm{otherwise.}
\end{array}\right.$$
Note that the action set includes the length $0$ vector, which we will denote by $\emptyset \equiv a^*_{1:0}$ and yields reward $R_{t+1,\emptyset} = 1$.

The reward process $\{R_{t+1}\}_{t \in \bZ_{\geq 0}}$ is exchangeable since $R_1=R_2=\cdots$.  As $\{R_{t+1}\}_{t \in \bZ_{\geq 0}}$ is determined by $a^*$, we define its prior distribution in terms of a prior distribution of $a^*$.  In particular, for each $k \in \bZ_{> 0}$, let the distribution $p_k(\cdot|a^*_{1:k-1})$ of $a^*_k$ conditioned on $a^*_{1:k-1}$ be geometric with mean $\tau$.  Then, let the prior probability assigned to $a^*_{1:K}$ be $\prod_{k=1}^K p_k(a^*_k|a^*_{1:k-1})$.
\label{xample:curricular}
\end{example}

We say that an agent is \emph{exploiting} at timestep $t$ if its chosen action $A_t$ is a previously selected action known to achieve highest reward, given history $H_t$. Otherwise, we say that an agent is \emph{exploring}. Example \ref{xample:curricular} offers a representative instance of how the presence of infinitely-many actions and unbounded rewards demands an infinite amount of exploration in order to synthesize optimal behavior. An initial thought may be to purely explore in perpetuity and continually uncover higher tiers of reward. Unfortunately, the cost structure associated with incorrect actions (that do not match the goal sequence $a^*$) is calibrated such that this policy is provably not optimal.

\begin{theorem}
\label{thm:explore}
    In Example \ref{xample:curricular}, an agent that always explores is never discounted-overtaking optimal. 
\end{theorem}

Upon further reflection, it is perhaps not terribly surprising that a strategy of pure exploration is sub-optimal in Example \ref{xample:curricular}, as is often the case for many sequential decision-making problems studied in the literature. However, unlike these latter commonly-studied problems, we may also obtain an analogous theoretical result establishing that any policy which ceases exploration in any time period is also provably not-optimal. This represents a more substantial departure from the traditional literature, where the presence of bounded rewards (even with infinitely-many actions) guarantees the existence of an optimal policy which eventually exploits with probability 1; we defer a review of this prior work to Section \ref{sec:disc}.

\begin{theorem}
\label{thm:exploit}
In Example \ref{xample:curricular}, an agent that stops exploring is never \gdot{}.
\end{theorem}

Naturally, if neither exploring nor exploiting with probability 1 yields an optimal strategy, an agent's only recourse is to randomize. Our next theoretical result formalizes this intuition.

\begin{theorem}
\label{thm:regret}
    In Example \ref{xample:curricular}, for all $T < \infty$, there exists a policy $\pi^*_T$ that is exploiting with non-zero probability $p^*_T \in (0,1)$ and exploring with non-zero probability $1-p^*_T$ at each timestep, such that, for all policy $\pi$, 
    $$
    \mathrm{Regret}^T_{\pi, \pi^*_T} \le 0. 
    $$
\end{theorem}

The precise probability with which an agent optimally balances its preference for exploration versus exploitation in each time period, $p^*_T$, is sensitive to the overall time horizon over which it aims to maximize reward. We offer the following conjecture to clarify how this probability behaves asymptotically as the agent engages with the full brunt of infinite exploration in a complex environment. 

\begin{conjecture} \label{conj:convergence}
    In Example \ref{xample:curricular}, as $T\to\infty$, $p_T^* \to \frac{\alpha+1}{\alpha+\tau}$. 
\end{conjecture}
Conjecture \ref{conj:convergence} posits that, as $T$ increases, the optimal policy randomizes between exploitation and exploration with exploiting probability $p_T^*$ approaching a non-zero threshold.  Intuitively, even as $T$ approaches infinity, the agent should never taper off its exploration probability.  To see why this conjecture holds, we provide a potential roadmap in Appendix \ref{sec:appendix2}. It is worth noting that there is a concrete instantiation of Example \ref{xample:curricular} for which we can offer a more precise, elegant characterization of the result in Conjecture \ref{conj:convergence}. Specifically, when $\alpha=2$; $\tau=4$; and $\gamma\ge 0.85$, $p_T^*\to\frac{\alpha+1}{\alpha+\tau} = 0.5$ and an optimal agent must perseverate with equiprobable exploration and exploitation for all time. 

\section{Discussion}
\label{sec:disc}

Two key facets of the complex environment studied in this work are the presence of infinitely-many actions and unbounded rewards. In this section, we begin with an overview of prior work, which largely focuses on the former condition in the absence of the latter, as well as a small handful of papers from outside the machine-learning literature which consider both conditions together. We conclude with a discussion of how one might begin to approach the design of practical agents for such complex environments and offer a simple computational experiment to corroborate our proposal.

\subsection{Prior Work}

Multi-armed bandit problems with infinitely-many actions available to the agent have been a topic of interest in the literature for decades. The earliest work by \citet{mallows1964some} demonstrates that reward distributions with bounded moments allows for the characterization of an optimal, non-stationary policy maximizing average reward. A similar non-stationary strategy is analyzed for regret minimization by \citet{yakowitz1991nonparametric}, who also rely on bounded higher moments of the reward distributions at each arm. \citet{banks1992denumerable} extend the classic machinery of Gittins' indices~\citep{gittins1974dynamic,gittins1979bandit,gittins1979dynamic} to bandit problems with a countably-infinite number of independent arms, each of which is assumed to have an associated distribution that yields uniformly bounded expected rewards. Identical structural assumptions are made explicitly, by \citet{lai1995machine,wang2008algorithms, carpentier2015simple,aziz2018pure,kalvit2020finite,de2021bandits,lai2022bandit,wang2022beyond,russo2022satisficing}, as well as implicitly, by \citet{herschkorn1996policies,berry1997bandit,chen2004note,chen2005note,hung2012optimal,bonald2013two,gong2023asymptotically}, where the latter all study the infinite-armed Bernoulli bandit. A related setting to the infinite-armed Bernoulli bandit is the so-called many-armed bandit setting where the number of actions is finite but considered large relative to the problem horizon~\citep{teytaud2007anytime,zhu2020regret}. Notably, all of the aforementioned papers do not entertain unbounded rewards alongside an infinitely-large action space.

\citet{agrawal1995continuum} studies bandit problems whose action space forms a subset of the real line $\bR$ under the assumptions of sub-Gaussian rewards and uniformly locally-Lipschitz continuous mean rewards; crucially, the latter assumption allows for a carefully-constructed collection of actions which adequately cover the action space to obtain an approximation of the expected reward function suitable for identifying near-optimal actions. Improvements of these results for the one-dimensional case as well as extensions to vector-valued action spaces of arbitrary dimension are studied by \citet{kleinberg2004nearly,auer2007improved,kleinberg2008multi,cope2009regret,bubeck2011x}. Aside from the boundedness of expected rewards implied by the sub-Gaussianity assumptions in some of these works, the more critical distinction is (again) in the use of uniform local-Lipschitz continuity, which serves as the key inductive bias to facilitate effective learning when there are infinitely-many actions. In contrast, the complex environment studied in this work presents a latent curricular structure that enables efficient learning despite the unboundedness of rewards. 

Unbounded rewards represent an important piece of the complex environment studied in this work, ensuring ample opportunity to dramatically improve the value of any current best decision known to an agent. Decision-making problems with unbounded rewards have long been a subject of study in philosophy~\citep{arntzenius2004bayesianism,goodsell2023decision}, though, to the best of the authors' knowledge, seem to not be a topic of study in the traditional bandit literature. A typical discussion point of such external papers to RL are the paradoxes that emerge among potentially optimal behaviors defined to maximize expected utility. Meanwhile, this work focuses on alternative performance criteria for which optimal behavior can be clearly defined, though may not be guaranteed to exist.

\subsection{Towards Practical Agent Design}

While we offer theoretical results which underscore the importance of randomization for a discounted-overtaking optimal policy in a complex environment, it is perhaps not immediately apparent how to go about the practical implementation of a computational agent that could learn or approximate this optimal behavior from interaction data. We anticipate that the concept of a \emph{learning target}~\citep{lu2023reinforcement} is essential to the design and practical implementation of such an agent. When, at any given time, optimal behavior is so complicated that it requires too much information to learn, it behooves the agent to have a mechanism for prioritizing some other modest corpus of information that, while capable of facilitating behavioral improvement, is itself insufficient to enable near-optimal performance; broadly speaking, a learning target is such a mechanism. As the agent gains competency through its prolonged interaction with the environment, one might envision that this learning target could adapt in kind to reflect updated knowledge and reorient exploration towards new, feasible discoveries. 


\begin{wrapfigure}{r}{.5\linewidth}
\vspace{-0.7cm}
\includegraphics[width=\linewidth]{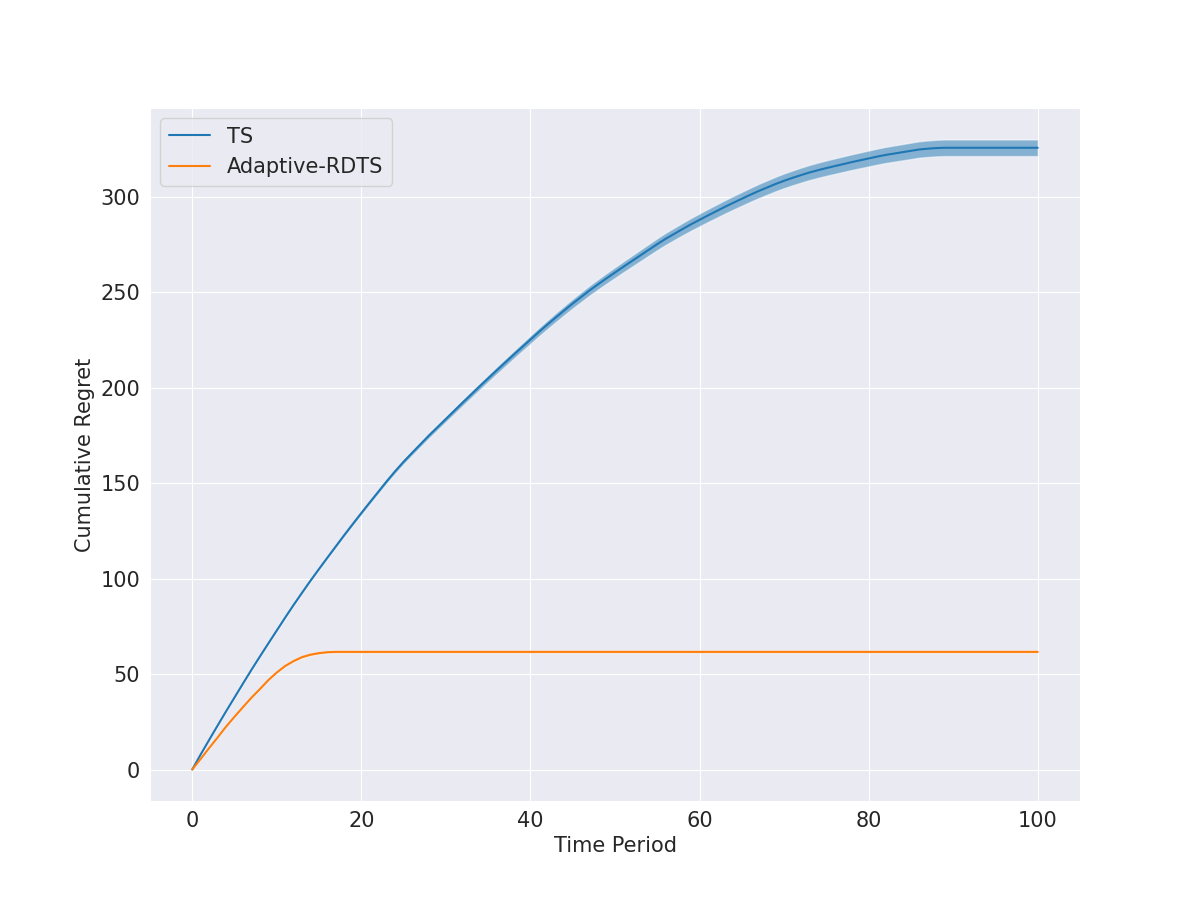}
\caption{Cumulative regret curve comparing Thompson Sampling and Rate-Distortion Thompson Sampling agents for learning the first two digits of $\pi$.}
\label{fig:cuml_regret}
\end{wrapfigure} 

A recent line of work~\citep{arumugam2021deciding,arumugam2021the,arumugam2022deciding} has studied the design, analysis, and implementation of decision-making agents endowed with the ability to compute such learning targets and autonomously decide what to learn. We strongly suspect that deciding what to learn and striking a desired trade-off between information requirements and performance is a critical capability for an agent coping with complex environments where eternal exploration is the only path to optimal behavior. As a preliminary empirical illustration of our hypothesis, we offer a computational result based on the $\pi$-guessing example of Section \ref{sec:intro}. For the sake of computational feasibility, we prune down the original $\pi$-guessing game and reduce the task to learning the first two digits of $\pi$, yielding a finite action set $\mc{A} = \{0, \ldots, 99\}$. We refer interested readers to Section \ref{sec:exp} of the appendix for more granular details of the computational experiment. 

A classic approach like Thompson Sampling~\citep{thompson1933likelihood,russo2018tutorial} (TS) invests exploratory effort only in those actions with non-zero probability of being optimal, thereby forgoing the opportunity to leverage the inherent curricular structure of the environment outline in Example \ref{xample:curricular}. Meanwhile, an adaptive variant of the Rate-Distortion Thompson Sampling (RDTS) agent introduced by \citet{arumugam2024bayesian} is able to compute a sequence of learning targets for scaffolding exploration around the identification of successive digits. Consequently, in the worst case, TS demands at most 90 time periods to identify the first two digits of $\pi$ whereas RDTS requires at most 20. Importantly, while the associated cumulative regret curves shown in Figure \ref{fig:cuml_regret} pertains to an environment lacking the key features of infinitely-many actions and unbounded rewards, it does illustrate how a learning target can be operationalized to modulate exploration in a manner resemblant of what would be needed for efficient learning in Example \ref{xample:curricular} and complex environments more broadly.

\section{Conclusion}
\label{sec:conc}

In this paper, we have engaged with a broadened treatment of the exploration challenge in sequential decision-making problems, departing from the traditional setting where optimal behavior tapers exploration over time. Instead, as an agent always has substantial opportunity for improvement, optimal behavior demands eternal exploration. A single, quintessential environment studied throughout this paper exemplifies this nuanced exploration problem through the combination of an infinitely-large action space and unbounded rewards. Our theoretical analysis clarifies the manner in which strategies of pure exploration and pure exploitation fall short of optimal performance, thereby necessitating the use of a randomization to preserve an agent's propensity to explore the world.  

We posit that our work offers a simple microcosm for the formal study of an exploration problem that manifests at a much grander scale across several real-world applications~\citep{shi2017world,toyama2021androidenv,stiennon2020learning,ouyang2022training,yao2022webshop,dwaracherla2024efficient}. Large language models (LLMs), across each individual user prompt, must contend with exploring the entire space of natural language responses to identify the best one~\citep{stiennon2020learning,ouyang2022training,dwaracherla2024efficient}. To impose an upper bound on rewards in LLMs (usually corresponding to human preference or utility scores) is to imply a known upper limit on human satisfaction across the entire space of natural language prompts. Instead, one might consider the possibility that such an upper limit on response utility does not exist and so, to any current best-known response, there always remains an opportunity to refine and improve. Our analysis suggests a natural exploration strategy for such settings where untested candidate responses are emitted in careful proportion to currently preferred responses over time. Similar scenarios are perhaps likely to emerge in environments of comparable scale, including those for learning desired behaviors on mobile devices~\citep{toyama2021androidenv} or the Internet itself~\citep{shi2017world,yao2022webshop}. Additionally, extending beyond the considerations of any single task, the lifelong or continual reinforcement learning setting has been a recent source of great interest to the community~\citep{ring1994continual,khetarpal2022towards,kumar2023continual,abel2024definition}. The inherent non-stationarity of the environment in continual reinforcement learning also beckons for a strategy of prolonged and enduring exploration, for which future theoretical analyses and agent-design principles may take inspiration from this work.




\bibliography{references}

\begin{thebibliography}{61}
\providecommand{\natexlab}[1]{#1}
\providecommand{\url}[1]{\texttt{#1}}
\expandafter\ifx\csname urlstyle\endcsname\relax
  \providecommand{\doi}[1]{doi: #1}\else
  \providecommand{\doi}{doi: \begingroup \urlstyle{rm}\Url}\fi

\bibitem[Abel et~al.(2024)Abel, Barreto, Van~Roy, Precup, van Hasselt, and
  Singh]{abel2024definition}
David Abel, Andr{\'e} Barreto, Benjamin Van~Roy, Doina Precup, Hado~P van
  Hasselt, and Satinder Singh.
\newblock {A Definition of Continual Reinforcement Learning}.
\newblock \emph{Advances in Neural Information Processing Systems}, 36, 2024.

\bibitem[Agrawal(1995)]{agrawal1995continuum}
Rajeev Agrawal.
\newblock {The Continuum-Armed Bandit Problem}.
\newblock \emph{SIAM Journal on Control and Optimization}, 33\penalty0
  (6):\penalty0 1926--1951, 1995.

\bibitem[Arimoto(1972)]{arimoto1972algorithm}
Suguru Arimoto.
\newblock {An Algorithm for Computing the Capacity of Arbitrary Discrete
  Memoryless Channels}.
\newblock \emph{IEEE Transactions on Information Theory}, 18\penalty0
  (1):\penalty0 14--20, 1972.

\bibitem[Arntzenius et~al.(2004)Arntzenius, Elga, and
  Hawthorne]{arntzenius2004bayesianism}
Frank Arntzenius, Adam Elga, and John Hawthorne.
\newblock {Bayesianism, Infinite Decisions, and Binding}.
\newblock \emph{Mind}, 113\penalty0 (450):\penalty0 251--283, 2004.

\bibitem[Arumugam \& Van~Roy(2021{\natexlab{a}})Arumugam and
  Van~Roy]{arumugam2021deciding}
Dilip Arumugam and Benjamin Van~Roy.
\newblock {Deciding What to Learn: A Rate-Distortion Approach}.
\newblock In \emph{International Conference on Machine Learning}, pp.\
  373--382. PMLR, 2021{\natexlab{a}}.

\bibitem[Arumugam \& Van~Roy(2021{\natexlab{b}})Arumugam and
  Van~Roy]{arumugam2021the}
Dilip Arumugam and Benjamin Van~Roy.
\newblock {The Value of Information When Deciding What to Learn}.
\newblock \emph{Advances in Neural Information Processing Systems},
  34:\penalty0 9816--9827, 2021{\natexlab{b}}.

\bibitem[Arumugam \& Van~Roy(2022)Arumugam and Van~Roy]{arumugam2022deciding}
Dilip Arumugam and Benjamin Van~Roy.
\newblock {Deciding What to Model: Value-Equivalent Sampling for Reinforcement
  Learning}.
\newblock In \emph{Advances in Neural Information Processing Systems},
  volume~35, 2022.

\bibitem[Arumugam et~al.(2024)Arumugam, Ho, Goodman, and
  Van~Roy]{arumugam2024bayesian}
Dilip Arumugam, Mark~K Ho, Noah~D Goodman, and Benjamin Van~Roy.
\newblock {Bayesian Reinforcement Learning with Limited Cognitive Load}.
\newblock \emph{Open Mind}, 8:\penalty0 395--438, 2024.

\bibitem[Auer et~al.(2007)Auer, Ortner, and Szepesv{\'a}ri]{auer2007improved}
Peter Auer, Ronald Ortner, and Csaba Szepesv{\'a}ri.
\newblock {Improved Rates for the Stochastic Continuum-Armed Bandit Problem}.
\newblock In \emph{International Conference on Computational Learning Theory},
  pp.\  454--468. Springer, 2007.

\bibitem[Aziz et~al.(2018)Aziz, Anderton, Kaufmann, and Aslam]{aziz2018pure}
Maryam Aziz, Jesse Anderton, Emilie Kaufmann, and Javed Aslam.
\newblock {Pure Exploration in Infinitely-Armed Bandit Models with
  Fixed-Confidence}.
\newblock In \emph{Algorithmic Learning Theory}, pp.\  3--24. PMLR, 2018.

\bibitem[Banks \& Sundaram(1992)Banks and Sundaram]{banks1992denumerable}
Jeffrey~S Banks and Rangarajan~K Sundaram.
\newblock {Denumerable-Armed Bandits}.
\newblock \emph{Econometrica: Journal of the Econometric Society}, pp.\
  1071--1096, 1992.

\bibitem[Berger(1971)]{berger1971rate}
Toby Berger.
\newblock \emph{{Rate Distortion Theory: A Mathematical Basis for Data
  Compression}}.
\newblock Prentice-Hall, 1971.

\bibitem[Berry et~al.(1997)Berry, Chen, Zame, Heath, and
  Shepp]{berry1997bandit}
Donald~A Berry, Robert~W Chen, Alan Zame, David~C Heath, and Larry~A Shepp.
\newblock {Bandit Problems with Infinitely Many Arms}.
\newblock \emph{The Annals of Statistics}, 25\penalty0 (5):\penalty0
  2103--2116, 1997.

\bibitem[Blahut(1972)]{blahut1972computation}
Richard Blahut.
\newblock {Computation of Channel Capacity and Rate-Distortion Functions}.
\newblock \emph{IEEE Transactions on Information Theory}, 18\penalty0
  (4):\penalty0 460--473, 1972.

\bibitem[Bonald \& Proutiere(2013)Bonald and Proutiere]{bonald2013two}
Thomas Bonald and Alexandre Proutiere.
\newblock {Two-Target Algorithms for Infinite-Armed Bandits with Bernoulli
  Rewards}.
\newblock \emph{Advances in Neural Information Processing Systems}, 26, 2013.

\bibitem[Bubeck et~al.(2011)Bubeck, Munos, Stoltz, and
  Szepesv{\'a}ri]{bubeck2011x}
S{\'e}bastien Bubeck, R{\'e}mi Munos, Gilles Stoltz, and Csaba Szepesv{\'a}ri.
\newblock X-armed bandits.
\newblock \emph{Journal of Machine Learning Research}, 12\penalty0 (5), 2011.

\bibitem[Carpentier \& Valko(2015)Carpentier and Valko]{carpentier2015simple}
Alexandra Carpentier and Michal Valko.
\newblock {Simple Regret for Infinitely Many Armed Bandits}.
\newblock In \emph{International Conference on Machine Learning}, pp.\
  1133--1141. PMLR, 2015.

\bibitem[Chen \& Lin(2004)Chen and Lin]{chen2004note}
Kung-Yu Chen and Chien-Tai Lin.
\newblock {A Note on Strategies for Bandit Problems with Infinitely Many Arms}.
\newblock \emph{Metrika}, 59:\penalty0 193--203, 2004.

\bibitem[Chen \& Lin(2005)Chen and Lin]{chen2005note}
Kung-Yu Chen and Chien-Tai Lin.
\newblock {A Note on Infinite-Armed Bernoulli Bandit Problems with Generalized
  Beta Prior Distributions}.
\newblock \emph{Statistical Papers}, 46\penalty0 (1):\penalty0 129--140, 2005.

\bibitem[Chiang \& Boyd(2004)Chiang and Boyd]{chiang2004geometric}
Mung Chiang and Stephen Boyd.
\newblock {Geometric Programming Duals of Channel Capacity and Rate
  Distortion}.
\newblock \emph{IEEE Transactions on Information Theory}, 50\penalty0
  (2):\penalty0 245--258, 2004.

\bibitem[Cope(2009)]{cope2009regret}
Eric~W Cope.
\newblock {Regret and Convergence Bounds for a Class of Continuum-Armed Bandit
  Problems}.
\newblock \emph{IEEE Transactions on Automatic Control}, 54\penalty0
  (6):\penalty0 1243--1253, 2009.

\bibitem[Cover \& Thomas(2012)Cover and Thomas]{cover2012elements}
Thomas~M Cover and Joy~A Thomas.
\newblock \emph{{Elements of Information Theory}}.
\newblock John Wiley \& Sons, 2012.

\bibitem[Csisz{\'a}r(1974)]{csiszar1974extremum}
Imre Csisz{\'a}r.
\newblock {On an Extremum Problem of Information Theory}.
\newblock \emph{Studia Scientiarum Mathematicarum Hungarica}, 9, 1974.

\bibitem[de~Finetti(1937)]{de1937foresight}
Bruno de~Finetti.
\newblock {Foresight: Its Logical Laws, its Subjective Sources}.
\newblock In \emph{Breakthroughs in Statistics: Foundations and Basic Theory},
  pp.\  134--174. Springer, 1937.

\bibitem[De~Heide et~al.(2021)De~Heide, Cheshire, M{\'e}nard, and
  Carpentier]{de2021bandits}
Rianne De~Heide, James Cheshire, Pierre M{\'e}nard, and Alexandra Carpentier.
\newblock {Bandits with Many Optimal Arms}.
\newblock \emph{Advances in Neural Information Processing Systems},
  34:\penalty0 22457--22469, 2021.

\bibitem[Diamond \& Boyd(2016)Diamond and Boyd]{diamond2016cvxpy}
Steven Diamond and Stephen Boyd.
\newblock {CVXPY: A Python-embedded Modeling Language for Convex Optimization}.
\newblock \emph{Journal of Machine Learning Research}, 17\penalty0
  (83):\penalty0 1--5, 2016.

\bibitem[Dwaracherla et~al.(2024)Dwaracherla, Asghari, Hao, and
  Van~Roy]{dwaracherla2024efficient}
Vikranth Dwaracherla, Seyed~Mohammad Asghari, Botao Hao, and Benjamin Van~Roy.
\newblock {Efficient Exploration for LLMs}.
\newblock \emph{arXiv preprint arXiv:2402.00396}, 2024.

\bibitem[Gittins(1974)]{gittins1974dynamic}
John Gittins.
\newblock {A Dynamic Allocation Index for the Sequential Design of
  Experiments}.
\newblock \emph{Progress in Statistics}, pp.\  241--266, 1974.

\bibitem[Gittins(1979)]{gittins1979bandit}
John Gittins.
\newblock {Bandit Processes and Dynamic Allocation Indices}.
\newblock \emph{Journal of the Royal Statistical Society Series B: Statistical
  Methodology}, 41\penalty0 (2):\penalty0 148--164, 1979.

\bibitem[Gittins \& Jones(1979)Gittins and Jones]{gittins1979dynamic}
John~C Gittins and David~M Jones.
\newblock {A Dynamic Allocation Index for the Discounted Multiarmed Bandit
  Problem}.
\newblock \emph{Biometrika}, 66\penalty0 (3):\penalty0 561--565, 1979.

\bibitem[Gong \& Sellke(2023)Gong and Sellke]{gong2023asymptotically}
Evelyn Xiao-Yue Gong and Mark Sellke.
\newblock {Asymptotically Optimal Quantile Pure Exploration for Infinite-Armed
  Bandits}.
\newblock \emph{Advances in Neural Information Processing Systems}, 36, 2023.

\bibitem[Goodsell(2023)]{goodsell2023decision}
Zachary Goodsell.
\newblock {Decision Theory Unbound}.
\newblock \emph{No{\^u}s}, 2023.

\bibitem[Herschkorn et~al.(1996)Herschkorn, Pekoez, and
  Ross]{herschkorn1996policies}
Stephen~J Herschkorn, Erol Pekoez, and Sheldon~M Ross.
\newblock {Policies Without Memory for the Infinite-Armed Bernoulli Bandit
  Under the Average-Reward Criterion}.
\newblock \emph{Probability in the Engineering and Informational Sciences},
  10\penalty0 (1):\penalty0 21--28, 1996.

\bibitem[Hung(2012)]{hung2012optimal}
Ying-Chao Hung.
\newblock {Optimal Bayesian Strategies for the Infinite-Armed Bernoulli
  Bandit}.
\newblock \emph{Journal of Statistical Planning and Inference}, 142\penalty0
  (1):\penalty0 86--94, 2012.

\bibitem[Kalvit \& Zeevi(2020)Kalvit and Zeevi]{kalvit2020finite}
Anand Kalvit and Assaf Zeevi.
\newblock {From Finite to Countable-Armed Bandits}.
\newblock \emph{Advances in Neural Information Processing Systems},
  33:\penalty0 8259--8269, 2020.

\bibitem[Khetarpal et~al.(2022)Khetarpal, Riemer, Rish, and
  Precup]{khetarpal2022towards}
Khimya Khetarpal, Matthew Riemer, Irina Rish, and Doina Precup.
\newblock {Towards Continual Reinforcement Learning: A Review and
  Perspectives}.
\newblock \emph{Journal of Artificial Intelligence Research}, 75:\penalty0
  1401--1476, 2022.

\bibitem[Kleinberg(2004)]{kleinberg2004nearly}
Robert Kleinberg.
\newblock {Nearly Tight Bounds for the Continuum-Armed Bandit Problem}.
\newblock \emph{Advances in Neural Information Processing Systems}, 17, 2004.

\bibitem[Kleinberg et~al.(2008)Kleinberg, Slivkins, and
  Upfal]{kleinberg2008multi}
Robert Kleinberg, Aleksandrs Slivkins, and Eli Upfal.
\newblock {Multi-Armed Bandits in Metric Spaces}.
\newblock In \emph{Proceedings of the Fortieth Annual ACM Symposium on Theory
  of Computing}, pp.\  681--690, 2008.

\bibitem[Kumar et~al.(2023)Kumar, Marklund, Rao, Zhu, Jeon, Liu, and
  Van~Roy]{kumar2023continual}
Saurabh Kumar, Henrik Marklund, Ashish Rao, Yifan Zhu, Hong~Jun Jeon, Yueyang
  Liu, and Benjamin Van~Roy.
\newblock {Continual Learning as Computationally Constrained Reinforcement
  Learning}.
\newblock \emph{arXiv preprint arXiv:2307.04345}, 2023.

\bibitem[Lai \& Yakowitz(1995)Lai and Yakowitz]{lai1995machine}
Tze-Leung Lai and Sidney Yakowitz.
\newblock {Machine Learning and Nonparametric Bandit Theory}.
\newblock \emph{IEEE Transactions on Automatic Control}, 40\penalty0
  (7):\penalty0 1199--1209, 1995.

\bibitem[Lai et~al.(2022)Lai, Sklar, and Xu]{lai2022bandit}
Tze~Leung Lai, Michael~Benjamin Sklar, and Huanzhong Xu.
\newblock {Bandit and Covariate Processes, with Finite or Non-Denumerable Set
  of Arms}.
\newblock \emph{Stochastic Processes and their Applications}, 150:\penalty0
  1222--1237, 2022.

\bibitem[Lattimore \& Szepesv{\'a}ri(2020)Lattimore and
  Szepesv{\'a}ri]{lattimore2020bandit}
Tor Lattimore and Csaba Szepesv{\'a}ri.
\newblock \emph{{Bandit Algorithms}}.
\newblock Cambridge University Press, 2020.

\bibitem[Lu et~al.(2023)Lu, Van~Roy, Dwaracherla, Ibrahimi, Osband, Wen,
  et~al.]{lu2023reinforcement}
Xiuyuan Lu, Benjamin Van~Roy, Vikranth Dwaracherla, Morteza Ibrahimi, Ian
  Osband, Zheng Wen, et~al.
\newblock {Reinforcement Learning, Bit by Bit}.
\newblock \emph{Foundations and Trends{\textregistered} in Machine Learning},
  16\penalty0 (6):\penalty0 733--865, 2023.

\bibitem[Mallows \& Robbins(1964)Mallows and Robbins]{mallows1964some}
CL~Mallows and Herbert Robbins.
\newblock {Some Problems of Optimal Sampling Strategy}.
\newblock \emph{Journal of Mathematical Analysis and Applications}, 8\penalty0
  (1):\penalty0 90--103, 1964.

\bibitem[Ouyang et~al.(2022)Ouyang, Wu, Jiang, Almeida, Wainwright, Mishkin,
  Zhang, Agarwal, Slama, Ray, et~al.]{ouyang2022training}
Long Ouyang, Jeff Wu, Xu~Jiang, Diogo Almeida, Carroll~L Wainwright, Pamela
  Mishkin, Chong Zhang, Sandhini Agarwal, Katarina Slama, Alex Ray, et~al.
\newblock {Training Language Models to Follow Instructions with Human
  Feedback}.
\newblock \emph{arXiv preprint arXiv:2203.02155}, 2022.

\bibitem[Puterman(1994)]{puterman1994markov}
Martin~L Puterman.
\newblock {Markov Decision Processes: Discrete Stochastic Dynamic Programming},
  1994.

\bibitem[Ring(1994)]{ring1994continual}
Mark~Bishop Ring.
\newblock \emph{{Continual Learning in Reinforcement Environments}}.
\newblock PhD thesis, The University of Texas at Austin, 1994.

\bibitem[Russo \& Van~Roy(2022)Russo and Van~Roy]{russo2022satisficing}
Daniel Russo and Benjamin Van~Roy.
\newblock {Satisficing in Time-Sensitive Bandit Learning}.
\newblock \emph{Mathematics of Operations Research}, 2022.

\bibitem[Russo et~al.(2018)Russo, Van~Roy, Kazerouni, Osband, and
  Wen]{russo2018tutorial}
Daniel~J Russo, Benjamin Van~Roy, Abbas Kazerouni, Ian Osband, and Zheng Wen.
\newblock {A Tutorial on Thompson Sampling}.
\newblock \emph{Foundations and Trends{\textregistered} in Machine Learning},
  11\penalty0 (1):\penalty0 1--96, 2018.

\bibitem[Shannon(1948)]{shannon1948mathematical}
Claude~E Shannon.
\newblock {A Mathematical Theory of Communication}.
\newblock \emph{The Bell System Technical Journal}, 27\penalty0 (3):\penalty0
  379--423, 1948.

\bibitem[Shannon(1959)]{shannon1959coding}
Claude~E. Shannon.
\newblock {Coding Theorems for a Discrete Source with a Fidelity Criterion}.
\newblock \emph{IRE Nat. Conv. Rec., March 1959}, 4:\penalty0 142--163, 1959.

\bibitem[Shi et~al.(2017)Shi, Karpathy, Fan, Hernandez, and
  Liang]{shi2017world}
Tianlin Shi, Andrej Karpathy, Linxi Fan, Jonathan Hernandez, and Percy Liang.
\newblock {World of Bits: An Open-Domain Platform for Web-Based Agents}.
\newblock In \emph{International Conference on Machine Learning}, pp.\
  3135--3144. PMLR, 2017.

\bibitem[Stiennon et~al.(2020)Stiennon, Ouyang, Wu, Ziegler, Lowe, Voss,
  Radford, Amodei, and Christiano]{stiennon2020learning}
Nisan Stiennon, Long Ouyang, Jeffrey Wu, Daniel Ziegler, Ryan Lowe, Chelsea
  Voss, Alec Radford, Dario Amodei, and Paul~F Christiano.
\newblock {Learning to Summarize with Human Feedback}.
\newblock \emph{Advances in Neural Information Processing Systems},
  33:\penalty0 3008--3021, 2020.

\bibitem[Teytaud et~al.(2007)Teytaud, Gelly, and Sebag]{teytaud2007anytime}
Olivier Teytaud, Sylvain Gelly, and Michele Sebag.
\newblock {Anytime Many-Armed Bandits}.
\newblock In \emph{CAP07}, 2007.

\bibitem[Thompson(1933)]{thompson1933likelihood}
William~R Thompson.
\newblock {On the Likelihood That One Unknown Probability Exceeds Another in
  View of the Evidence of Two Samples}.
\newblock \emph{Biometrika}, 25\penalty0 (3/4):\penalty0 285--294, 1933.

\bibitem[Toyama et~al.(2021)Toyama, Hamel, Gergely, Comanici, Glaese, Ahmed,
  Jackson, Mourad, and Precup]{toyama2021androidenv}
Daniel Toyama, Philippe Hamel, Anita Gergely, Gheorghe Comanici, Amelia Glaese,
  Zafarali Ahmed, Tyler Jackson, Shibl Mourad, and Doina Precup.
\newblock {AndroidEnv: A Reinforcement Learning Platform for Android}.
\newblock \emph{arXiv preprint arXiv:2105.13231}, 2021.

\bibitem[Wang et~al.(2022)Wang, Baharav, Han, Jiao, and Tse]{wang2022beyond}
Yifei Wang, Tavor Baharav, Yanjun Han, Jiantao Jiao, and David Tse.
\newblock {Beyond the Best: Distribution Functional Estimation in
  Infinite-Armed Bandits}.
\newblock \emph{Advances in Neural Information Processing Systems},
  35:\penalty0 9262--9273, 2022.

\bibitem[Wang et~al.(2008)Wang, Audibert, and Munos]{wang2008algorithms}
Yizao Wang, Jean-Yves Audibert, and R{\'e}mi Munos.
\newblock {Algorithms for Infinitely Many-Armed Bandits}.
\newblock \emph{Advances in Neural Information Processing Systems}, 21, 2008.

\bibitem[Yakowitz \& Lowe(1991)Yakowitz and Lowe]{yakowitz1991nonparametric}
Sid Yakowitz and Wing Lowe.
\newblock {Nonparametric Bandit Methods}.
\newblock \emph{Annals of Operations Research}, 28:\penalty0 297--312, 1991.

\bibitem[Yao et~al.(2022)Yao, Chen, Yang, and Narasimhan]{yao2022webshop}
Shunyu Yao, Howard Chen, John Yang, and Karthik~R Narasimhan.
\newblock {WebShop: Towards Scalable Real-World Web Interaction with Grounded
  Language Agents}.
\newblock In \emph{Advances in Neural Information Processing Systems}, 2022.

\bibitem[Zhu \& Nowak(2020)Zhu and Nowak]{zhu2020regret}
Yinglun Zhu and Robert Nowak.
\newblock {On Regret with Multiple Best Arms}.
\newblock \emph{Advances in Neural Information Processing Systems},
  33:\penalty0 9050--9060, 2020.

\end{thebibliography}
\bibliographystyle{rlc}

\appendix

\section{Computational Experiment}
\label{sec:exp}

In this section, we provide additional details on the concluding computational experiment of Section \ref{sec:disc}. The environment can be seen as a restricted version of Example \ref{xample:curricular} where the action set $\mc{A} = \{0,1,\ldots,99\}$ consists of all two-digit sequences and the agent aims to learn the first two digits of $\pi$ with $\alpha = 2$. 

Given the agent's current (posterior) beliefs about the underlying environment $\bP(\theta \in \cdot \mid H_t)$, a Thompson Sampling agent proceeds via the probability-matching principle to select an action $A_t$ such that $\bP(A_t = a \mid H_t) = \bP(A^\star = a \mid H_t)$, where $A^\star \in \argmax\limits_{a \in \mc{A}} \bE\left[R_{1, a} \mid \theta \right].$ Broadly speaking, one might consider an alternative learning target $\chi \in \mc{A}$ such that an agent may employ a variant of Thompson Sampling by probability matching with respect to $\chi$: $\bP(A_t = a \mid H_t) = \bP(\chi = a \mid H_t)$. 

A line of work~\citep{arumugam2021deciding,arumugam2021the,arumugam2024bayesian} studies how to compute such a learning target via information theory~\citep{shannon1948mathematical,cover2012elements}. More specifically, when targeting an optimal action $A^\star$, the mutual information between the environment and target $\bI_t(\theta; A^\star)$ given the current (random) history $H_t$ quantifies the amount of information an agent must obtain through prudent exploration in order to identify optimal behavior. In the context of this work, a complex environment is one for which this amount of information is near-infinite or intractably large $\bI_t(\theta;A^\star) \uparrow \infty$ across all time periods. Consequently, an agent may instead find it fruitful to orient exploration around an alternative target $\chi$ that is easier to learn, in the sense that $\bI_t(\theta; \chi) \leq \bI_t(\theta; A^\star)$. Of course, as the agent still aims to be productive with respect to the task at hand and optimize reward, this should be done carefully so as to incur bounded expected regret in each time period: $\bE_t\left[R_{t,A^\star} - R_{t, \chi}\right] \leq D$, for some threshold $D \in \bR_{\geq 0}$.

Striking a desired balance between information and utility is a hallmark characteristic of lossy compression problems studied by the information theory community within the sub-area of rate-distortion theory~\citep{shannon1959coding,berger1971rate}. The fundamental limit for the lossy compression faced by an agent is given by the rate-distortion function $$\mc{R}_t(D) = \inf\limits_{\widetilde{A}} \bI_t(\theta; \widetilde{A}) \text{ such that } \bE_t\left[\left(R_{t,A^\star} - R_{t, \widetilde{A}}\right)^2\right] \leq D.$$

The Rate-Distortion Thompson Sampling (RDTS) algorithm of \citet{arumugam2024bayesian} provides a theoretical analysis outlining the benefits of computing and probability matching with respect to the target $\widetilde{A}_t$ that achieves the rate-distortion limit in each time period. While their study pertains to a fixed distortion threshold $D \in \bR_{\geq 0}$, our computational experiments employ an adaptive version where the dynamic threshold $D_t$ is chosen to ensure the agent focuses its efforts on identifying the digits of $\pi$ in sequence, rather than pursuing a guess for all digits at once like Thompson Sampling (corresponding to $D_t = 0$ for all time periods). While previous work~\citep{arumugam2021deciding} has avoided dealing directly with the rate-distortion function computationally by appealing to the classic Blahut-Arimoto algorithm~\citep{blahut1972computation,arimoto1972algorithm}, our experiment leverages the fact that the rate-distortion function constitutes a convex optimization problem~\citep{csiszar1974extremum,chiang2004geometric} which can be solved in each time period via CVXPY~\citep{diamond2016cvxpy}.

\section{Analysis}
\label{sec:appendix2}

In this section, we provide all proofs for theoretical results presented in the main paper.

\subsection{Proof of Theorem \ref{thm:exploit}}

\begin{proof}
For ease of exposition, we refer to each component in an action tuple $(a_1, \ldots, a_k)$ as one \emph{digit}. The analogy is made for the illustrative $\pi$-guessing game in Section \ref{sec:intro}.  A history $H_t = (A_0, R_1, A_1, \ldots, A_t, R_{t+1})$ can be completely characterized by an agent state $S_t$ made up of digits tried and failed so far as well as $a_k^*$ known so far.  Selecting digits that the agent has tried and failed before is clearly suboptimal, as those digits incur a cost yet offer no new information.  Therefore, it suffices to restrict our attention to (stationary) policies with states $S_t$ that do not try digits already attempted and failed.  

By the temporal symmetric structure of Example \ref{xample:curricular}, if $a_{1:k}^* \in H_t$ and $a^*_{1:k+1} \notin H_t$, a (stationary) policy at time $t$ must sample from two actions: 1) exploit $a_{1:k}^*$ or 2) explore the next digit.  

We consider two classes of policies:
\begin{enumerate}
    \item $\pi_N$: for each $N \in \mathbb{Z}_{>0}$, $\pi_N$ explores to identify $a_{0:N}^*$ sequentially, then exploits $a_{0:N}^*$. 
    \begin{center}
    \begin{tikzpicture}
    \node[state] (a0) {$a^*_0$};
    \node[state, right of=a0] (a1) {$a_{0:1}^*$};
    \node[state, right of=a1] (a2) {$a_{0:2}^*$};
    \node[state, right of=a2] (a3) {$a_{0:3}^*$};
    \node[right of=a3] (a4) {$\cdots$};
    \node[state, right of=a4] (aN) {$a_{0:N}^*$};
    \draw (a0) edge node {$\mu_1$} (a1);
    \draw (a1) edge node {$\mu_2$} (a2);
    \draw (a2) edge node {$\mu_3$} (a3);
    \draw (a3) edge node {$\mu_4$} (a4);
    \draw (a4) edge node {$\mu_N$} (aN);
    \draw (aN) edge[loop above] node {} (aN);
    \end{tikzpicture}
    \end{center}
    \item $\pi^\mathrm{explore}$: $\pi^\mathrm{explore}$ always explores sequentially. 
    \begin{center}
    \begin{tikzpicture}
    \node[state] (a0) {$a^*_0$};
    \node[state, right of=a0] (a1) {$a_{0:1}^*$};
    \node[state, right of=a1] (a2) {$a_{0:2}^*$};
    \node[state, right of=a2] (a3) {$a_{0:3}^*$};
    \node[right of=a3] (a4) {$\cdots$};
    \draw (a0) edge node {$\mu_1$} (a1);
    \draw (a1) edge node {$\mu_2$} (a2);
    \draw (a2) edge node {$\mu_3$} (a3);
    \draw (a3) edge node {$\mu_4$} (a4);
    \end{tikzpicture}
    \end{center}
\end{enumerate}

We recursively define stopping times for the exploration of each digit under the always exploring agent $\pi^\mathrm{explore}$.  Let $\mu_0 = 1$ and $\mu_k = \min\{t > 0: A_{t + \sum_{j=0}^{k-1} \mu_j} = a_{1:k}^*, A_i \sim \pi^\mathrm{explore} \}$.  In other words, $\mu_k$ denotes the time it takes for $\pi^\mathrm{explore}$ to discover the $k$-th digit given that it knows the first $k-1$ digits.  The agent's prior implies that $\mu_k$ is i.i.d. geometric with mean $\tau$.  An useful quantity is the expected discount factor at $\mu_k$:
\begin{align*}
    \E\left[\gamma^{\mu_k}\right] &= \sum\limits_{j=1}^\infty \bP(\mu_k = j) \gamma^j = \sum\limits_{j=1}^\infty \left(1 - \frac{1}{\tau}\right)^{j-1} \frac{1}{\tau} \gamma^j = \frac{\gamma}{(1-\gamma)\tau + \gamma}.
\end{align*}
We consider two cases: $\gamma < 1$ and $\gamma = 1$.  When $\gamma < 1$, the expected reward at time $\sum_{j=0}^k \mu_j+1$ is
$$
\E \left[ \gamma^{\sum_{j=0}^k \mu_j-1} \alpha^k \right] = \frac{\alpha}{(1-\gamma)\tau+\gamma}\left(\frac{\alpha\gamma}{(1-\gamma)\tau+\gamma}\right)^{k-1},
$$
and the expected cost accumulated while exploring for $a_k^*$ satisfies
$$
\E \left[\sum_{i=\sum_{j=0}^{k-1}\mu_j+1}^{\sum_{j=0}^{k}
\mu_j -1} \gamma^{i-1}\frac{\alpha+1}{\tau-1} \alpha^{k-1}\right] = \frac{\alpha+1}{(1-\gamma)\tau+\gamma}\left(\frac{\alpha\gamma}{(1-\gamma)\tau+\gamma}\right)^{k-1} = \frac{\alpha+1}{\alpha}\E \left[ \gamma^{\sum_{j=0}^k \mu_j-1} \alpha^k \right].
$$
Assuming that $T$ is large enough such that $T\gg \sum_{j=0}^N \mu_j$ for fixed $N$ almost surely, we calculate the returns over a horizon $T$. 
\begin{align*}
    V_{\pi_N}^{\gamma,T} &= \E\left[\sum_{k=0}^{N-1}\gamma^{\sum_{j=0}^k \mu_j-1}\alpha^k - \sum_{k=1}^N \frac{\alpha+1}{\alpha} \gamma^{\sum_{j=0}^k \mu_j-1} \alpha^k + \sum_{i=\sum_{j=0}^N \mu_j}^T \gamma^{i-1}\alpha^N \right] \\
    &= 1 - \sum_{k=1}^{N-1} \frac{1}{(1-\gamma)\tau+\gamma} \left(\frac{\alpha\gamma}{(1-\gamma)\tau+\gamma}\right)^{k-1} - \frac{\alpha+1}{(1-\gamma)\tau+\gamma}\left(\frac{\alpha\gamma}{(1-\gamma)\tau+\gamma}\right)^{N-1} \\
    &\quad + \alpha^N\frac{\left(\frac{\gamma}{(1-\gamma)\tau+\gamma}\right)^{N} - \gamma^{T+1}}{\gamma(1-\gamma)} \\
    &= 1 - \frac{1}{(1-\gamma)\tau+(1-\alpha)\gamma} + \left(\frac{1}{(1-\gamma)\tau+(1-\alpha)\gamma} + \frac{1}{1-\gamma}\right)\left(\frac{\gamma}{(1-\gamma)\tau+\gamma}\right)^{N} - \frac{\alpha^N}{1-\gamma}\gamma^T.
\end{align*}
For $N_1 < N_2$, we have that 
\begin{align*}
    V_{\pi_{N_2}}^{\gamma,T} - V_{\pi_{N_1}}^{\gamma,T} &= \left(\frac{1}{(1-\gamma)\tau+(1-\alpha)\gamma} + \frac{1}{1-\gamma}\right)\left[\left(\frac{\alpha\gamma}{(1-\gamma)\tau+\gamma}\right)^{N_2} - \left(\frac{\alpha\gamma}{(1-\gamma)\tau+\gamma}\right)^{N_1}\right] \\
    &\quad - \frac{\alpha^{N_2} - \alpha^{N_1}}{1-\gamma}\gamma^T \\
    &\xrightarrow{T\to\infty} \left(\frac{1}{(1-\gamma)\tau+(1-\alpha)\gamma} + \frac{1}{1-\gamma}\right)\left[\left(\frac{\alpha\gamma}{(1-\gamma)\tau+\gamma}\right)^{N_2} - \left(\frac{\alpha\gamma}{(1-\gamma)\tau+\gamma}\right)^{N_1}\right] \\
    &> 0.
\end{align*}

This implies that a policy can always improve by exploring more digits before committing.  Therefore, a policy that stops exploring is never \gdot{}.

When $\gamma=1$, the expected cost accumulated while exploring for $a_k^*$ satisfies:
$$
\E \left[\sum_{i=\sum_{j=0}^{k-1}\mu_j+1}^{\sum_{j=0}^{k}
\mu_j -1} \frac{\alpha+1}{\tau-1} \alpha^{k-1}\right] = (\alpha+1)\alpha^{k-1}.
$$
Assuming that $T$ is large enough such that $T \gg \sum_{j=0}^N \mu_j$ for fixed $N$ almost surely, we calculate the returns over a horizon $T$:
\begin{align*}
    V^{1,T}_{\pi_N} &= \E\left[ \sum_{k=0}^{N-1} \alpha^k - \sum_{k=1}^{N} \sum_{i=\sum_{j=0}^{k-1}\mu_j+1}^{\sum_{j=0}^{k}
\mu_j -1} \frac{\alpha+1}{\tau-1} \alpha^{k-1} + \sum_{i=\sum_{j=0}^N \mu_j}^{T} \alpha^N\right] \\
&= \E\left[ \sum_{k=0}^{N-1} \alpha^k - \sum_{k=1}^{N} (\alpha+1)\alpha^{k-1} + (T - \sum_{j=0}^N \mu_j + 1)\alpha^N\right] \\
&= -\alpha\frac{\alpha^N-1}{\alpha-1} + (T - N\tau)\alpha^N. 
\end{align*}
For $N_1 < N_2$, we have that 
\begin{align*}
    \liminf_{T\to\infty} (V_{\pi_{N_2}}^{1,T} - V_{\pi_{N_1}}^{1,T}) &> 0.
\end{align*}
This again implies that a policy can always improve by exploring more digits before committing.  Therefore, a policy that stops exploring is never \gdot{}.  Moreover,
$$
\mathrm{Regret}_{\pi_{N_2}, \pi_{N_1}}^T \ge \mathcal{O}(T).
$$
\end{proof}

\subsection{Proof of Theorem \ref{thm:explore}}

\begin{proof}
We consider two cases: $\gamma = 1$ and $\gamma < 1$.  When $\gamma<1$, the return of the exploring agent over a finite horizon $T$ can be computed as
\begin{align*}
    V^{\gamma,T}_{\pi^\mathrm{explore}} &= \E \Bigg[ \sum_{N=1}^\infty \left(\sum_{k=0}^N \gamma^{\sum_{j=0}^k \mu_j-1} \alpha^k - \sum_{k=1}^N \frac{\alpha+1}{\alpha}\gamma^{\sum_{j=0}^k \mu_j-1} \alpha^k - \sum_{i=\sum_{j=0}^N \mu_j+1}^T \gamma^{i-1} \frac{\alpha+1}{\tau-1}\alpha^N \right)\\
    &\quad \cdot \1\{\sum_{j=0}^N \mu_j \le T < \sum_{j=0}^{N+1}\mu_j\}\Bigg] \\
    &\le \E \left[ \sum_{N=1}^\infty \left(\sum_{k=0}^N \gamma^{\sum_{j=0}^k \mu_j-1} \alpha^k - \sum_{k=1}^N \frac{\alpha+1}{\alpha}\gamma^{\sum_{j=0}^k \mu_j-1} \alpha^k \right) \cdot \1\{\sum_{j=0}^N \mu_j \le T < \sum_{j=0}^{N+1}\mu_j\}\right] \\
    &= \E\left[ \sum_{N=1}^\infty \left(1 - \frac{1}{(1-\gamma)\tau+\gamma} \sum_{k=1}^N \left(\frac{\alpha\gamma}{(1-\gamma)\tau+\gamma}\right)^{k-1}\right)  \cdot \1\{\sum_{j=0}^N \mu_j \le T < \sum_{j=0}^{N+1}\mu_j\} \right].
\end{align*}
Comparing $\pi_1$ and $\pi^\mathrm{explore}$, 
\begin{align*}
    V_{\pi_1}^{1,T} - V_{\pi^\mathrm{explore}}^{1,T} &= \sum_{N=1}^\infty \sum_{k=2}^N \left(\frac{\alpha\gamma}{(1-\gamma)\tau+\gamma}\right)^{k-1} \cdot \1\{\sum_{j=0}^N \mu_j \le T < \sum_{j=0}^{N+1}\mu_j\} + \frac{1}{1-\gamma} \left(\frac{\alpha\gamma}{(1-\gamma)\tau+\gamma}\right) \\
    &\quad - \frac{\gamma^T}{1-\gamma}\alpha \\
    &\xrightarrow{T\to\infty} \left(\frac{\alpha\gamma}{(1-\gamma)\tau+\gamma}\right) \frac{1}{1 - \frac{\alpha\gamma}{(1-\gamma)\tau+\gamma}} + \frac{1}{1-\gamma} \left(\frac{\alpha\gamma}{(1-\gamma)\tau+\gamma}\right) \\
    &= \left(\frac{\alpha\gamma}{(1-\gamma)\tau+\gamma}\right) \left(\frac{(1-\gamma)\tau+\gamma}{(1-\gamma)\tau+(1-\alpha)\gamma} + \frac{1}{1-\gamma}\right).
\end{align*}
Since $\tau < \frac{\gamma(\alpha-1)}{2(1-\gamma)}$, one can check that 
$$
\frac{(1-\gamma)\tau+\gamma}{(1-\gamma)\tau+(1-\alpha)\gamma} + \frac{1}{1-\gamma} > 0
$$
and thus $\liminf_{T\to\infty}(V_{\pi_1}^{1,T} - V_{\pi^\mathrm{explore}}^{1,T}) > 0$.  This implies that the exploring agent using $\pi^\mathrm{explore}$ is never \gdot{}.  

When $\gamma=1$, the return of the exploring agent over a finite horizon $T$ can be computed as
    \begin{align*}
        V^{1,T}_{\pi^\mathrm{explore}} &= \E\left[ \sum_{N=1}^\infty \left( \sum_{k=0}^{N} \alpha^k - \sum_{k=1}^{N} \sum_{i=\sum_{j=0}^{k-1}\mu_j+1}^{\sum_{j=0}^{k}
\mu_j -1} \frac{\alpha+1}{\tau-1} \alpha^{k-1} - (T - \sum_{j=0}^N \mu_j)\frac{\alpha+1}{\tau-1} \alpha^N\right)\cdot \1\{\sum_{j=0}^{N}\mu_j \le T < \sum_{j=0}^{N+1} \mu_j\} \right] \\
&\le \E\left[ \sum_{N=1}^\infty \left( \sum_{k=0}^{N} \alpha^k - \sum_{k=1}^{N} \sum_{i=\sum_{j=0}^{k-1}\mu_j+1}^{\sum_{j=0}^{k}
\mu_j -1} \frac{\alpha+1}{\tau-1} \alpha^{k-1} \right)\cdot \1\{\sum_{j=0}^{N}\mu_j \le T < \sum_{j=0}^{N+1} \mu_j\} \right] \\
&= \E\left[ \sum_{N=1}^\infty \left(\alpha^N - \alpha\frac{\alpha^N-1}{\alpha-1} \right) \cdot \1\{\sum_{j=0}^{N}\mu_j \le T < \sum_{j=0}^{N+1} \mu_j\} \right] \\
&\le \E\left[ \sum_{N=1}^\infty \left(\alpha^N - \alpha^{N} \right) \cdot \1\{\sum_{j=0}^{N}\mu_j \le T < \sum_{j=0}^{N+1} \mu_j\} \right] = 0.
    \end{align*}
Thus, comparing $\pi_1$ and $\pi^\mathrm{explore}$,
$$
\liminf_{T\to\infty} (V_{\pi_1}^{1,T} - V_{\pi^\mathrm{explore}}^{1,T}) > 0.
$$
This implies that the exploring agent using $\pi^\mathrm{explore}$ is never \gdot{}.  Moreover,
$$
\mathrm{Regret}_{\pi_1, \pi^\mathrm{explore}}^{T} \ge \mathcal{O}(T).
$$
\end{proof}

\subsection{Proof of Theorem \ref{thm:regret}}

\begin{proof}
    We consider three additional classes of policies to those mentioned in the proof of Theorem \ref{thm:exploit}:
    \begin{enumerate}
    \item Stochastic policies $\pi^p$: for $p\in[0,1)$, $\pi^p$ exploits the best known action with probability $p$ and explores the next digit with probability $1-p$ at each time.  Note that $\pi^0 = \pi^\mathrm{explore}$.
    \item Non-curricular policies $\pi_N'$: for each $N \in \mathbb{Z}_{>0}$, $\pi'_N$ explores to identify $a_{1:N}^*$ directly, then exploits $a_{1:N}^*$. 
    \begin{center}
    \begin{tikzpicture}
    \node[state] (a0) {$a^*_0$};
    \node[state, right of=a0] (aN) {$a_{0:N}^*$};
    \draw (a0) edge node {$\mu'_N$} (a1);
    \draw (aN) edge[loop above] node {} (aN);
    \end{tikzpicture}
    \end{center}
    \item Non-stationary policies $\pi^\mathrm{NS}_{m}$: for $m\in\mathbb{R}_{\ge 0}$, after each $a_{1:k}^*$ is discovered, $\pi^\mathrm{NS}_{m}$ exploits $a_{1:k}^*$ for $m$ times before exploring the next digit. 
    \begin{center}
    \begin{tikzpicture}
    \node[state] (a0) {$a^*_0$};
    \node[state, right of=a0] (a1) {$a_{0:1}^*$};
    \node[state, right of=a1] (a2) {$a_{0:2}^*$};
    \node[state, right of=a2] (a3) {$a_{0:3}^*$};
    \node[right of=a3] (a4) {$\cdots$};
    \draw (a0) edge node {$\mu_1$} (a1);
    \draw (a1) edge node {$\mu_2$} (a2);
    \draw (a2) edge node {$\mu_3$} (a3);
    \draw (a3) edge node {$\mu_4$} (a4);
    \draw (a0) edge[loop above] node {$m$} (a0);
    \draw (a1) edge[loop above] node {$m$} (a1);
    \draw (a2) edge[loop above] node {$m$} (a2);
    \draw (a3) edge[loop above] node {$m$} (a3);
    \end{tikzpicture}
    \end{center}
    \end{enumerate}
    First we note a one-to-one mapping between the set of stochastic policies \{$\pi^p\}_{p\in[0,1)}$ and non-stationary policies $\{\pi^\mathrm{NS}_{m}\}_{m\in\mathbb{R}_{\ge 0}}$.  Indeed, when $p = \frac{m+1}{m+\tau}$, each $a^*_{1:k}$ is selected $m+1$ times on average. 

    For non-curricular policies, under the agent's prior belief, the optimal strategy of guessing is to guess all the combinations of $(a_1,\ldots,a_N)$ where $\sum_{i=1}^N a_i = N$, then those where $\sum_{i=1}^N a_i = N+1$, and so on.  We take $\pi'_N$ to be such a policy and define a stopping time $\mu_N' = \min\{t>0 | A_t = a^*_{1:N}, A_i \sim \pi'_N\}$.  We calculate the returns for $\pi'_N$ and $\pi^p$ as follows. 
\begin{align}
    V^{1,T}_{\pi_m^\mathrm{NS}} &= \sum_{n=1}^\infty \E \Bigg[ \1\{\sum_{j=0}^n \mu_j + nm \le T\} \left((m+1)\alpha^{n-1} - (\mu_n-1)\frac{\alpha+1}{\tau-1}\alpha^{n-1}\right) \nonumber\\
    &\quad + \1\{\sum_{j=1}^{n}\mu_j + nm \le T \le \sum_{j=1}^{n} \mu_j + (n+1)m\} \left(T - \sum_{j=1}^{n}\mu_j - nm + 1\right)\alpha^n \nonumber\\
    &\quad + \1\{\sum_{j=1}^{n}\mu_j + (n+1)m < T < \sum_{j=1}^{n+1} \mu_j + (n+1)m\} \left((m+1)\alpha^n - (T - \sum_{j=1}^{n} \mu_j - (n+1)m)\frac{\alpha+1}{\tau-1}\alpha^n\right) \Bigg]. \label{eqn:value-nonstationary} \\
    V^{1,T}_{\pi'_N} &= \E \left[ - \frac{\alpha+1}{\tau-1}\alpha^{N-1} \mu'_N + (T - \mu'_N + 1)\alpha^N)\right] \\
    &= - \E\left[\mu'_N\right]\frac{\alpha+1}{\tau-1}\alpha^{N-1} + \E\left[T - \mu'_N + 1\right]\alpha^N.
\end{align}
Note that if $m>T$, $V^{1,T}_{\pi_m^\mathrm{NS}} = 0$.  Hence, it suffices to consider $m\in[0,T]$.  Since $V^{1,T}_{\pi_m^\mathrm{NS}}$ is continuous in $m$ on $[0,T]$, by the extreme value theorem, there exists an $m^*\in[0,T]$ such that $V^{1,T}_{\pi_{m^*}^\mathrm{NS}} = \max_{m\in[0,T]} V^{1,T}_{\pi_{m}^\mathrm{NS}}$.  Taking $p_T^* = \frac{m^*}{m^*+1}$, we have
$$
\mathrm{Regret}_{\pi,\pi_T^*} \le 0. 
$$
Finally, we prove that an agent is better off following the curriculum, i.e., guessing one digit at a time in order.  Recall that
$$
V^{1,T}_{\pi_N} = -\alpha\frac{\alpha^N-1}{\alpha-1} + (T - N\tau)\alpha^N. 
$$
For $j \in\mathbb{Z}_{\ge 0}$, define $s_j = \sum_{i=1}^j \binom{N+i-2}{N-1}$, then
\begin{align*}
    \E\left[\mu'_N\right] &= \sum_{n=N}^\infty \sum_{\ell=s_{n-N}+1}^{s_{n-N+1}} \ell \left(1 - \frac{1}{\lambda}\right)^{n-N}\left(\frac{1}{\lambda}\right)^N \\
    &= \sum_{n=N}^\infty \frac{(s_{n-N}+s_{n-N+1}+1)(s_{n-N+1}-s_{n-N})}{2}\left(1 - \frac{1}{\lambda}\right)^{n-N} \left(\frac{1}{\lambda}\right)^N \\
    &= \frac{1}{2}\sum_{n=N}^\infty \left[2\sum_{i=1}^{n-N} \binom{N+i-2}{N-1} + \binom{n-1}{N-1} + 1\right]\binom{n-1}{N-1} \left(1 - \frac{1}{\lambda}\right)^{n-N} \left(\frac{1}{\lambda}\right)^N \\
    &\gg N\tau + 1. 
\end{align*}
Thus, $\E[T - \mu'_N + 1] < T - N\tau$ and 
$$
\E[\mu'_N]\frac{\alpha+1}{\tau-1}\alpha^{N-1} > (N\tau+1)\frac{\alpha+1}{\tau-1}\alpha^{N-1} > \alpha\frac{\alpha^N-1}{\alpha-1}
$$
for sufficiently large $N$.  Thus, there exists an $N_0 \in \mathbb{Z}_{>0}$ such that for all $N>N_0$, $V_{\pi'_N}^{1,T} < V_{\pi_N}^{1,T}$.  

\end{proof}

\subsection{Proof Roadmap for Conjecture \ref{conj:convergence}}

\begin{proof}[Proof roadmap]
    Recall the expression for $V^{1,T}_{\pi_m^\mathrm{NS}}$ in Equation \eqref{eqn:value-nonstationary}.  Consider the first summand in the expected value.  We let
    \begin{align*}
        f_n(m;T) &= \E \left[ \1\{\sum_{j=0}^n \mu_j + nm \le T\} \left((m+1)\alpha^{n-1} - (\mu_n-1)\frac{\alpha+1}{\tau-1}\alpha^{n-1}\right)\right].
    \end{align*}
 To decouple the dependence between the indicator random variable $\1\{\sum_{j=0}^n \mu_j + nm \le T\}$ and the multiplier $\left((m+1)\alpha^{n-1} - (\mu_n-1)\frac{\alpha+1}{\tau-1}\alpha^{n-1}\right)$, we define an independent copy $\tilde{\mu}_n$ of $\mu_n$ and analyze 
 \begin{align*}
 \tilde{f}_n(m;T) &= \E \left[ \1\{\sum_{j=0}^{n-1} \mu_j + \tilde{\mu}_n + nm \le T\} \left((m+1)\alpha^{n-1} - (\mu_n-1)\frac{\alpha+1}{\tau-1}\alpha^{n-1}\right)\right] \\
 &= \Pr\left(\sum_{j=0}^{n-1} \mu_j + \tilde{\mu}_n + nm \le T\right) \E\left[(m+1)\alpha^{n-1} - (\mu_n-1)\frac{\alpha+1}{\tau-1}\alpha^{n-1}\right] \\
 &= \Pr\left(\sum_{j=0}^{n-1} \mu_j + \tilde{\mu}_n + nm \le T\right) (m-\alpha)\alpha^{n-1}. 
\end{align*}
We may then proceed to lower bound $\Pr\left(\sum_{j=0}^{n-1} \mu_j + \tilde{\mu}_n + nm \le T\right)$ using Kolmogorov's inequality.  An upper bound on $\tilde{f}_n(m;T)$ can be obtained by taking each $\mu_j = 1$ in the indicator.  Finally, we account for the difference $f_n(m;T) - \tilde{f}_n(m;T)$.  A similar analysis can be carried out for the second and third summand in Equation \eqref{eqn:value-nonstationary}. 
 Optimizing the upper and lower bounds of $V^{1,T}_{\pi_m^\mathrm{NS}}$ over $m$ gives two sequences $\overline{m}^*_T$ and $\hat{m}^*_T$, respectively, which both converge to $\alpha$.  Thus, the corresponding exploitation probabilities should converge to $\frac{\alpha+1}{\alpha+\tau}$. 
\end{proof}

\end{document}